\documentclass{article}

\PassOptionsToPackage{round,authoryear}{natbib}
\usepackage[preprint]{neurips_2026}
\makeatletter
\renewcommand{\@noticestring}{Preprint.}
\makeatother

\usepackage[utf8]{inputenc}
\usepackage[T1]{fontenc}
\usepackage{microtype}
\usepackage{xcolor}
\usepackage{graphicx}
\usepackage{booktabs}
\usepackage{amsmath, amssymb}
\usepackage{float}
\usepackage{placeins}
\usepackage{hyperref}
\usepackage{caption}
\hypersetup{colorlinks=true, linkcolor=blue!50!black, citecolor=blue!50!black,
            urlcolor=blue!50!black}

\title{When Cosine Similarity Fails to Reflect\\
Linearly Accessible Structure in Dialogue Models}

\author{%
  Yu Sun \\ \texttt{ysun1@linkedin.com} \And
  Mengyin Lu \\ \texttt{melu@linkedin.com} \And
  Cong Feng \\ \texttt{cofeng@linkedin.com} \AND
  Guangming Lu \\ \texttt{glu@linkedin.com} \And
  Huimin Han \\ \texttt{huhan@linkedin.com}%
}

\date{}

\begin{document}
\maketitle

\begin{abstract}

Cosine similarity is widely used to analyze transformer representations, implicitly assuming that similarity reflects task-relevant structure. We study when this assumption fails in dialogue-conditioned large language models. Across three $7$--$8$B chat-tuned models, ambient cosine similarity substantially underestimates linearly decodable persona structure on the same hidden states; numerically, linear probe AUC is in the $0.73$--$0.97$ range while cosine kNN is in the $0.56$--$0.77$ range on a $30$-class task. A low-dimensional supervised subspace recovers much of this gap, whereas a matched-rank PCA subspace does not and in some cases degrades performance. This mismatch is regime-dependent: it is absent in single-sentence sentiment classification (SST-5), and a matched-cardinality control rules out attribute cardinality as a confound. The gap does not systematically increase across dialogue turns, and the task-aligned subspace remains stable over time. However, two of three models violate a pre-registered within-subspace separability invariance criterion ($|\Delta\text{AUC}| \le 0.03$), and one model violates a pre-registered turn-invariance criterion ($|\Delta L| \le 0.05$). These results show that cosine similarity can fail to reflect task-aligned structure in dialogue representations even when that structure is linearly accessible.
\end{abstract}

\section{Introduction}
\label{sec:intro}

Cosine similarity is widely used to analyze transformer representations---from persona-vector retrieval to representation-quality evaluation and steering verification---implicitly assuming that similarity reflects task-relevant structure. We study when this assumption fails: \textbf{when does similarity in representation space fail to reflect linearly decodable structure in dialogue model representations?}

\paragraph{Main finding.} Across three $7$--$8$B chat-tuned models (Llama-3.1-8B-Instruct, Qwen-3-8B, Mistral-7B-Instruct-v0.3) on a $30$-persona multi-turn dialogue setup, linear probes recover substantially more persona structure than cosine similarity on the same hidden states (Table~\ref{tab:headline}). A $10$-dimensional supervised subspace closes most of the gap (lift $+0.10$ to $+0.14$ over cosine kNN), while a matched-rank PCA subspace does not (lift $\le 0$; supervised$-$PCA gap $0.14$--$0.17$, bootstrap CIs strictly positive). The separation survives a $20\times$ cross-family activation-norm difference.

\paragraph{Multi-axis characterization.} We separate three diagnostics of this gap (not standalone phenomena). Along the turn axis, the supervised$-$PCA gap does not systematically increase across the dialogue, and the cumulative-pool-anchored supervised subspace $V^\star$ remains stable (projection-energy max-minus-min $\le 0.7$ percentage points). However, two of three models violate a pre-registered within-subspace separability invariance criterion ($|\Delta\text{AUC}| \le 0.03$), and one model (Llama) violates a pre-registered turn-invariance criterion ($|L_{T_{15}}-L_{T_1}| \le 0.05$); across four predefined persona attribute axes we observe no consistent turn-dependent structure ($|\Delta| \le 0.07$, mixed signs).

\paragraph{Positioning.} Recent work documents persona-conditioned behavioral collapse \citep{chameleon2026,stable2026,heterogeneity2026,systematic_persona2026}; we focus on \emph{representation-level} geometry rather than behavioral outputs and make no causal claim linking representation structure to behavioral diversity. We do not propose a principle, theorem, or causal mechanism for the cosine--linear mismatch; this paper documents an empirical regularity within the studied setting and characterizes its scope.

\paragraph{Relation to classical results.} The separation between task-aligned directions and variance-dominant directions is well known in classical multivariate statistics (e.g., Fisher's Linear Discriminant vs.\ PCA), and high-dimensional probing routinely observes that linear probes outperform distance-based methods. Our contribution is not this separation per se, but its systematic manifestation in dialogue-conditioned LLM representations across three $7$--$8$B chat-tuned models, its absence at matched cardinality in single-sentence settings, and the resulting implications for similarity-based analyses of transformer hidden states (persona-vector retrieval, representation-quality evaluation, steering verification) that take ambient cosine as a default geometry.

\paragraph{Contributions.} (i) a regime-dependent cosine--linear gap across three dialogue-conditioned $7$--$8$B settings; (ii) a supervised--vs--PCA subspace contrast separating task-aligned from variance-aligned structure; (iii) a multi-axis characterization of the boundary conditions under which this gap holds.

\bigskip
\noindent\fbox{\parbox{0.97\linewidth}{%
\textbf{Empirical regularity.} Cosine-based neighborhood geometry systematically diverges from linearly recoverable structure in dialogue-conditioned representations across three $7$--$8$B chat-tuned models, while disappearing in matched-cardinality single-sentence sentiment.
}}

\section{Related Work}
\label{sec:related}

\paragraph{Linear probing and the linear representation hypothesis.}
The linear-probing tradition \citep{hewitt2019designing,tenney2019bert,belinkov2022probing} establishes that many attributes are linearly readable from intermediate transformer representations. The linear representation hypothesis \citep{park2024linear} formalizes the idea that semantic concepts correspond to directions in residual-stream space. Our work extends this view by showing that the linear directions are not aligned with the variance-dominant directions of the ambient representation; ambient-space similarity therefore systematically underestimates the structure that linear probes recover.

\paragraph{Persona representation, steering, and subspace methods.}
A growing line of work studies persona at the behavior level---collapse and homogenization across long interactions \citep{chameleon2026,stable2026,heterogeneity2026,systematic_persona2026,emergent_misalign2026,pragmatic2026}---or constructs steering apparatus assuming persona directions are linearly recoverable: persona vectors and CAA \citep{panickssery2024caa,anthropic2025personavectors,arditi2024refusal,tan2024analyzing,engels2024notall}, and subspace methods such as EpiPersona \citep{epipersona2026}, which projects from prompt and context features onto a low-dimensional persona space, and \citet{wang2025persona}, which proposes a dual-head probing architecture for persona steering on Qwen2.5-0.5B. Our analysis is upstream of intervention and orthogonal to behavior: we measure whether the recoverable persona structure is reflected in ambient similarity geometry on internal hidden states across $7$--$8$B-class models, providing measurement-level support for projecting onto a task-aligned linear subspace rather than relying on ambient cosine.

\paragraph{Anisotropy and probing baselines.}
Anisotropy of transformer hidden states \citep{ethayarajh2019contextual,gao2019representation,timkey2021bark,mu2017abtt} produces well-known issues for similarity-based analyses; the cosine--linear mismatch we observe is preserved under all standard anisotropy corrections (Appendix~\ref{app:controls}). The probing literature \citep{belinkov2022probing,hewitt2019designing} routinely observes that linear probes outperform distance-based methods; our contribution beyond this is the regime-dependence (gap appears in dialogue-conditioned but not single-sentence settings at matched cardinality) and the supervised-vs-PCA contrast (Appendix~\ref{app:regime-robustness}).

\paragraph{Concept erasure.}
A line of work \citep{bolukbasi2016man,ravfogel2020null,belrose2023leace} uses supervised subspace projection to \emph{remove} attribute information; we project \emph{onto} the discriminative subspace for measurement rather than intervention (Appendix~\ref{app:erasure}). Detailed differentiation from concurrent persona-geometry, persona-vector, and EpiPersona work is in Appendix~\ref{app:related-detail}.

\section{Methods}
\label{sec:methods}

We treat representation structure operationally via a class of decoders (linear probes, supervised-subspace projection, PCA-subspace projection, ambient cosine kNN), and compare how they characterize the same hidden states.

\subsection{Dataset}
\label{sec:data}

Our dataset comprises $30$ manually-written persona characters spanning four axes (formality $2$-class; expertise $2$-class; emotion $5$-class; domain $5$-class), organized around a $5 \times 5$ emotion-domain design with formality and expertise varied across instances; we did not optimize selection for probe separability (full corpus and \texttt{adversarial\_group} annotation in Appendix~\ref{app:personas}). Each persona is paired with $5$ cross-context topics and rolled out over $15$ assistant turns at $N=4$ trajectories per (persona, topic) cell. While $N=4$ per cell is small, headline results pool across personas and topics ($600$ trajectories per model); fold-level $\sigma \le 0.006$ on all metrics from the $5{\times}2$ audit (\S\ref{sec:metrics}).
 Three models are evaluated: Llama-3.1-8B-Instruct ($32$ decoder layers), Qwen-3-8B (HuggingFace path \texttt{Qwen/Qwen3-8B}, $36$ decoder layers) \citep{qwen3}, and Mistral-7B-Instruct-v0.3 ($32$ decoder layers), yielding $600$ trajectories per model. We use the chat/instruct variants for all three model families.

We capture residual-stream activations at six approximately depth-matched layers per model (Llama: $\{8,12,16,20,24,28\}$; Qwen: $\{9,14,18,22,27,31\}$; Mistral: $\{8,12,16,20,24,28\}$) using forward hooks on the post-block output of each chosen decoder layer. Hidden states are extracted from the last prefill token (the activation conditioned on the full chat history, before generation begins) and stored as bf16-cast-to-float32 vectors. Headline-layer results are reported at the deepest probed layer per model: Llama L$28$, Qwen L$31$, Mistral L$28$.

\subsection{Probes and metrics}
\label{sec:metrics}
For a set of activations $X$ at a fixed (model, layer, slice), we compute four metrics under group-by-trajectory cross-validation. Headline regime-contrast results (\S\ref{sec:res-stage1}--\S\ref{sec:res-regime}) use $2$-fold group-by-trajectory CV (each fold holds half the trajectories per cell). We report variance separately via a matched $5\!\times\!2$ grouped CV audit at the same cells (Table~\ref{tab:headline}; point estimates within $|\Delta| \le 0.005$ of the $2$-fold values, $\sigma \le 0.006$ across all metrics; full numbers in Appendix~\ref{app:controls}). Turn-sliced and axis-sliced sub-experiments (\S\ref{sec:res-turn}--\S\ref{sec:res-axis}) use $5\!\times\!2$ grouped CV ($10$ folds) directly.%

\begin{itemize}
\item \textbf{Linear probe AUC.} $30$-class multinomial logistic regression with per-fold StandardScaler and L2 regularization $C{=}1.0$, lbfgs solver, max-iter $1{,}000$. Macro one-vs-rest AUC. We fix $C\!=\!1.0$ for all experiments and treat linear probe AUC as an operational reference ceiling under a single probe configuration rather than as an optimized estimate; the cosine--linear gap is therefore a comparison under fixed probe regularization.
\item \textbf{Ambient cosine kNN AUC.} \texttt{KNeighborsClassifier} with cosine distance, \texttt{weights="distance"}, and $k\!\in\!\{5, 10, 20\}$. We use cosine as the canonical default for transformer hidden-state similarity analyses; Euclidean kNN with the same activations gives comparable AUC (Appendix~\ref{app:controls}).
\item \textbf{Supervised subspace cosine kNN AUC.} Project activations onto the top-$k_\text{sub}$ right singular vectors of the train-fold linear-probe coefficient matrix $W_\text{train}$, then run cosine kNN. We use $k_\text{sub}\!=\!10$ for the full $30$-persona task.
\item \textbf{PCA-defined subspace cosine kNN AUC.} Same as supervised but project onto the top-$k_\text{sub}$ PCA components of train-fold scaled features.
\end{itemize}

Projection bases are fit strictly on training folds under group-by-trajectory cross-validation, preventing label leakage; details in Appendix~\ref{app:controls}. We sweep $k_\text{sub}\!\in\!\{1, 2, 5, 10, 20, 50\}$ and report the supervised lift $L\!=\!\text{AUC}_\text{sup}-\text{AUC}_\text{amb}$ and supervised-minus-PCA gap $S\!-\!P$, with trajectory-level bootstrap $95\%$ CIs for turn-sliced analyses.

\subsection{Controlled settings}
\label{sec:methods-controlled}
We evaluate three controlled settings on the same probe family (full setup in Appendix~\ref{app:methods-detail}). We refer to the persona-emotion-vs-SST-5 comparison as our \emph{matched-cardinality regime contrast}, used consistently in main text and appendix.

\begin{itemize}
\item \textbf{Per-axis probes} per persona axis at $k_\text{sub}\!=\!\max(\text{n\_classes}\!-\!1, 1)$, where the $5$-class persona-emotion case matches SST-5's cardinality.
\item \textbf{SST-5 sentiment} \citep{socher2013recursive,socher2013parsing} with $1{,}000$ sentences ($200$ per class) presented as single user-role messages (no system prompt, no rollout) under stratified $5$-fold CV.
\item \textbf{Turn slicing} per turn $k\!\in\!\{1, 3, 5, 7, 9, 11, 13, 15\}$, sample size held constant per turn, with $k_\text{knn}\!\in\!\{5, 10, 20\}$ as a density-stability check.
\end{itemize}

\subsection{Cumulative-anchored subspace}
\label{sec:methods-cumulative}
To distinguish lift invariance under per-turn-fitted bases from subspace invariance, we anchor the supervised subspace on the cumulative pool of all turn activations and project per-turn test data onto this fixed basis. Concretely: pool all $T_1$--$T_{15}$ activations per model ($N_\text{pool}\!=\!4{,}800$); under group-by-trajectory $5\!\times\!2$ cross-validation, fit logistic regression on the pooled-turns train fold to obtain $W_\text{train}$, take the top-$10$ right singular vectors via SVD to obtain $V^\star_\text{train} \in \mathbb{R}^{10 \times d}$. For each turn $k$ and test fold, we project test-fold activations at turn $k$ onto $V^\star_\text{train}$ and compute (i) projection energy $\|X_k V^{\star\top}\|_F^2 / \|X_k\|_F^2$ and (ii) within-subspace cosine kNN AUC.

\paragraph{Controls and pre-registration.}
Standard anisotropy corrections (L2 normalization, mean centering, all-but-the-top projection), layer sweep, and per-layer/per-turn norm fingerprints are reported in Appendix~\ref{app:controls}. Pre-registered decision rules (regime-gap thresholds, turn invariance, subspace support and separability invariance) are in Appendix~\ref{app:prereg}. We study the dialogue-conditioned regime as a jointly instantiated operational unit; controls test invariance over the joint distribution rather than isolating individual factors (Appendix~\ref{app:control-mapping}).

\section{Results}
\label{sec:results}

We organize results around a single primary empirical observation: a systematic separation between cosine-similarity neighborhood structure and linearly decodable structure in dialogue-conditioned persona representations (\S\ref{sec:res-stage1}--\S\ref{sec:res-regime}). Across architectures, supervised-subspace projection yields consistent lift ($+0.105$ to $+0.138$), while PCA-defined subspaces of matched dimensionality do not, producing a supervised-minus-PCA gap of $0.142$--$0.172$. This separation is observed across three model families and persists under multiple controls, suggesting it is not specific to a particular architecture or evaluation choice. Subsequent analyses---turn axis (\S\ref{sec:res-turn}), subspace stability (\S\ref{sec:res-subspace}), and attribute-axis decomposition (\S\ref{sec:res-axis})---characterize the scope and boundary conditions of this observation rather than independent findings.

\begin{figure}[!htbp]
\centering
\includegraphics[width=0.95\linewidth]{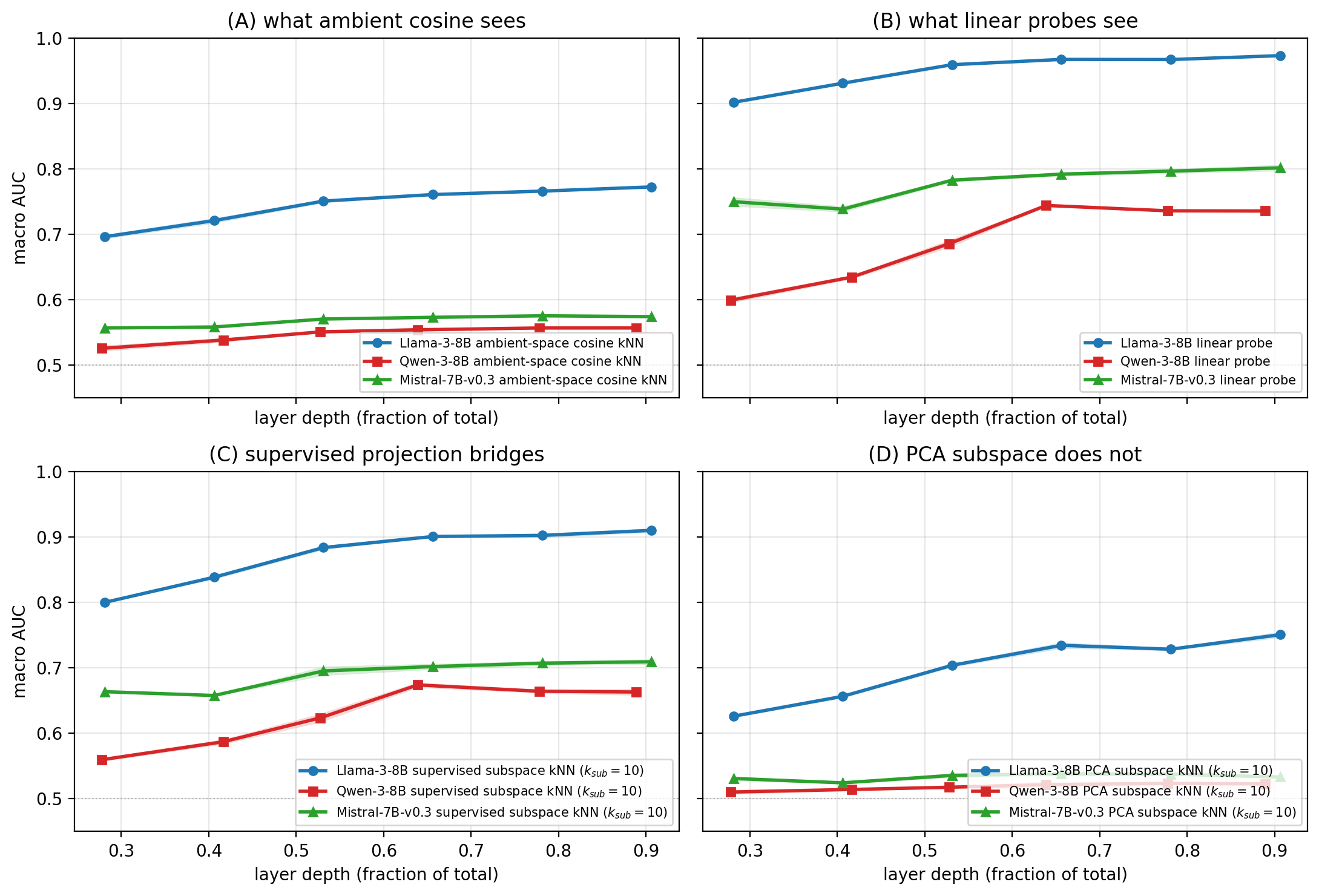}
\caption{\textbf{Cosine similarity yields substantially lower AUC than linear probes, while task-aligned subspaces recover most of the gap.} \textbf{(A)} Cosine kNN AUC per layer is low and family-dependent. \textbf{(B)} The same hidden states are linearly decodable with substantial cross-family ceiling variation ($0.733$--$0.972$). \textbf{(C)} A $10$-dimensional supervised subspace lifts cosine kNN by $+0.10$ to $+0.14$. \textbf{(D)} PCA subspaces of matched dimensionality remain at or below ambient. All panels: group-by-trajectory $5\!\times\!2$ CV at $k_\text{sub}\!=\!10$, persona-30-class; shaded bands are $\pm\sigma$ across folds. Same protocol as Figs.~\ref{fig:per-axis-regime}--\ref{fig:subspace-stability}.}
\label{fig:regime-contrast}
\end{figure}

\subsection{Stage 1: Representation mismatch --- cosine underestimates linearly decodable structure}
\label{sec:res-stage1}

Ambient-space cosine kNN yields substantially lower AUC than linear probes on all three models. At the deepest probed layer, cosine kNN with $k\!=\!10$ achieves macro AUC $0.771$ on Llama-3.1-8B (L$28$), $0.556$ on Qwen-3-8B (L$31$), and $0.574$ on Mistral-7B-Instruct-v0.3 (L$28$). These compare to a shuffled-label baseline of $0.500$ and to linear-probe AUCs reported in Stage 2 ($0.972$ / $0.733$ / $0.804$). The gap is preserved across all six probed layers in all three models.

The pattern is robust to standard anisotropy corrections (L2 normalization, mean centering, all-but-the-top projection) and persists despite a $20\times$ difference in mean activation norm across families (Appendix~\ref{app:controls}).

\subsection{Stage 2: Linearity of structure --- linear probes recover the signal, with cross-family variation in ceiling}
\label{sec:res-stage2}

The same hidden states are linearly decodable, with substantial cross-family variation in ceiling. Macro AUC of $30$-way logistic regression at the headline layer is $0.972$ (Llama L$28$), $0.733$ (Qwen L$31$), and $0.804$ (Mistral L$28$); the shuffled-label control yields $0.500$ on all three. The linear-vs-ambient gap is $0.20$ / $0.18$ / $0.23$: persona is therefore linearly encoded in all three architectures, but the ambient cosine metric does not reflect it.

\subsection{Stage 3: Task-aligned subspace recovery --- supervised projection lifts cosine kNN}
\label{sec:res-stage3}

Projecting onto a $10$-dimensional supervised subspace---the top-$10$ right singular vectors of the train-fold linear-probe coefficient matrix---substantially improves cosine kNN performance. At the headline layer, supervised-subspace kNN AUC is $0.909$ / $0.661$ / $0.709$ (Llama / Qwen / Mistral), corresponding to lifts of $+0.138$ / $+0.105$ / $+0.135$ over ambient cosine. The lift saturates near $k_\text{sub}\!=\!10$--$20$ and is preserved across all six probed layers (Appendix~\ref{app:controls}), indicating that a low-dimensional task-aligned subspace (10 dimensions out of $4096$) captures most of the linearly decodable signal under cosine similarity.

\subsection{Stage 4: Variance-aligned control --- PCA-defined subspaces do not recover comparable structure}
\label{sec:res-stage4}

A PCA-defined subspace of matched dimensionality serves as a variance-aligned control. At $k\!=\!10$, PCA-subspace kNN AUC is $0.749$ / $0.519$ / $0.537$, corresponding to lifts of $-0.022$ / $-0.037$ / $-0.037$ relative to ambient cosine. PCA therefore consistently underperforms the ambient representation across all three architectures and at every probed layer (no PCA lift exceeds zero for any tested $k_\text{sub} \in \{1, 2, 5, 10, 20, 50\}$).%

The supervised-minus-PCA gap is $\mathbf{0.160}$ / $\mathbf{0.142}$ / $\mathbf{0.172}$ (Table~\ref{tab:headline}), with bootstrap $95\%$ CIs strictly positive. This gap reflects two effects: a gain from task-aligned projection (supervised lift over ambient: $+0.138$ / $+0.105$ / $+0.135$, \S\ref{sec:res-stage3}) and a degradation under variance-aligned projection (PCA lift below ambient: $-0.022$ / $-0.037$ / $-0.037$). The supervised lift is therefore a property of task-alignment rather than of low-dimensional projection per se; variance-aligned projections can suppress the relevant geometric signal relative to the ambient space.

\begin{table}[!htbp]
\centering
\small
\caption{Headline-layer results on the $30$-class persona task: Llama-3.1-8B layer $28$, Qwen-3-8B layer $31$, Mistral-7B-Instruct-v0.3 layer $28$. Point estimates report macro one-vs-rest AUC under group-by-trajectory $2$-fold CV ($N\!=\!4$ trajectories per (persona, topic) cell); each fold holds out half the trajectories to maximize independence across correlated rollouts. The $\pm\sigma$ values give fold-level standard deviations from a matched $5\!\times\!2$ grouped CV audit at the same cells, which yields point estimates within $|\Delta| \le 0.005$ of the $2$-fold values shown (Appendix~\ref{app:controls}). The supervised-vs-PCA gap is consistent across all three architectures and bootstrap $95\%$ CIs are strictly positive.}
\label{tab:headline}
\begin{tabular}{lccc}
\toprule
Method                                          & Llama-3.1-8B (L$28$) & Qwen-3-8B (L$31$) & Mistral-7B (L$28$) \\
\midrule
Shuffled-label baseline                         & $0.500$ & $0.500$ & $0.500$ \\
Linear probe                                    & $0.972 \pm 0.001$ & $0.733 \pm 0.003$ & $0.804 \pm 0.005$ \\
Ambient cosine kNN ($k\!=\!10$)                 & $0.771 \pm 0.001$ & $0.556 \pm 0.001$ & $0.574 \pm 0.001$ \\
Supervised subspace kNN ($k_\text{sub}\!=\!10$) & $\mathbf{0.909 \pm 0.003}$ & $\mathbf{0.661 \pm 0.005}$ & $\mathbf{0.709 \pm 0.004}$ \\
PCA subspace kNN ($k_\text{sub}\!=\!10$)        & $0.749 \pm 0.005$ & $0.519 \pm 0.002$ & $0.537 \pm 0.002$ \\
\midrule
Supervised lift over ambient                    & $\mathbf{+0.138 \pm 0.003}$ & $\mathbf{+0.105 \pm 0.005}$ & $\mathbf{+0.135 \pm 0.005}$ \\
PCA lift over ambient                           & $-0.022 \pm 0.005$ & $-0.037 \pm 0.002$ & $-0.037 \pm 0.002$ \\
Supervised$-$PCA gap at $k_\text{sub}\!=\!10$   & $\mathbf{0.160 \pm 0.002}$ & $\mathbf{0.142 \pm 0.004}$ & $\mathbf{0.172 \pm 0.005}$ \\
\bottomrule
\end{tabular}
\end{table}

\subsection{Regime contrast under matched cardinality}
\label{sec:res-regime}

We compare persona-emotion ($5$-class) with SST-5 sentiment under matched label cardinality and matched subspace rank ($k_\text{sub}\!=\!5$). This controls for attribute cardinality and probe capacity, enabling a direct comparison of supervised-subspace behavior across two commonly studied settings: multi-turn dialogue and single-sentence classification (setup details in Appendix~\ref{app:methods-peraxis}).

The two settings are not fully matched. Persona-emotion uses a larger effective sample size (approximately $5\!\times$ more training points under matched CV), exhibits greater linear--ambient headroom prior to projection ($\sim\!0.18$--$0.23$ vs.\ $0.029$--$0.091$ on SST-5), and contains within-trajectory dependence across rollouts, whereas SST-5 consists of independent sentences. We therefore do not interpret this comparison as isolating a single causal factor.%

Despite these differences, the qualitative pattern is consistent across all three model families. In persona-emotion, supervised projection produces a substantial positive lift ($+0.095$ Llama, $+0.130$ Qwen, $+0.154$ Mistral) and a large supervised$-$PCA gap ($+0.246$ / $+0.194$ / $+0.213$), comparable to the $30$-class setting. In SST-5 at matched $k_\text{sub}\!=\!4$, supervised projection yields $-0.024$ Llama, $+0.013$ Qwen, $-0.063$ Mistral, with the supervised$-$PCA gap correspondingly collapsing. The contrast is therefore not only in magnitude but in direction: two of three SST-5 conditions exhibit negative supervised lift, whereas all three persona-emotion conditions are positive and substantial. A confound that scaled primarily with sample size or headroom would predict a reduced positive lift on SST-5, not a sign reversal.

\begin{figure}[!htbp]
\centering
\includegraphics[width=0.95\linewidth]{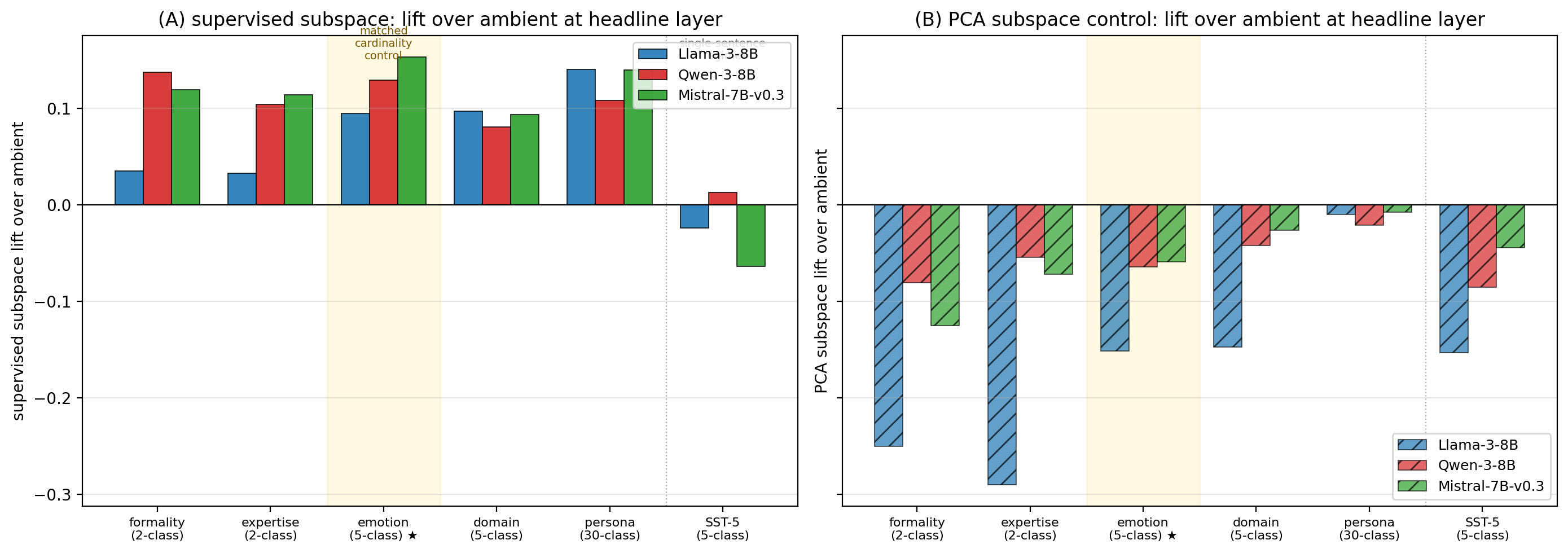}
\caption{\textbf{The cosine--linear misalignment is consistent across persona axes but absent in sentiment, even at matched cardinality.} \textbf{(A)} Supervised lift over ambient cosine kNN at the headline layer, per axis. Bars left of the dotted separator are persona axes (formality / expertise $2$-class, emotion / domain $5$-class, full $30$-class); right is SST-5 sentiment. Yellow shading: matched-cardinality control ($5$-class persona-emotion vs $5$-class SST-5). \textbf{(B)} Matched-rank PCA subspaces (hatched) yield zero or negative lift across all axes and models. The yellow vs rightmost-column comparison in (A) rules out attribute cardinality alone.}
\label{fig:per-axis-regime}
\end{figure}

The sign reversal between persona-emotion and SST-5 is inconsistent with explanations based solely on attribute cardinality, headroom, or sample size (a sample-size-matched robustness check preserves the contrast; Appendix~\ref{app:regime-robustness}). Within the dialogue-conditioned setting we study, ambient cosine similarity is therefore not a reliable proxy for task-aligned persona structure, and a low-dimensional supervised projection closes most of the gap where a matched-rank variance-aligned projection cannot; we do not extend this to untested regimes (\S\ref{sec:limitations}). Figure~\ref{fig:per-axis-regime} extends the comparison: supervised lift is positive across all $2$/$5$/$30$-class persona axes for all three architectures and zero or negative on SST-5, while PCA produces no lift in any setting.

\subsection{Turn axis: no systematic accumulation across turns}
\label{sec:res-turn}

\begin{figure}[!htbp]
\centering
\includegraphics[width=0.95\linewidth]{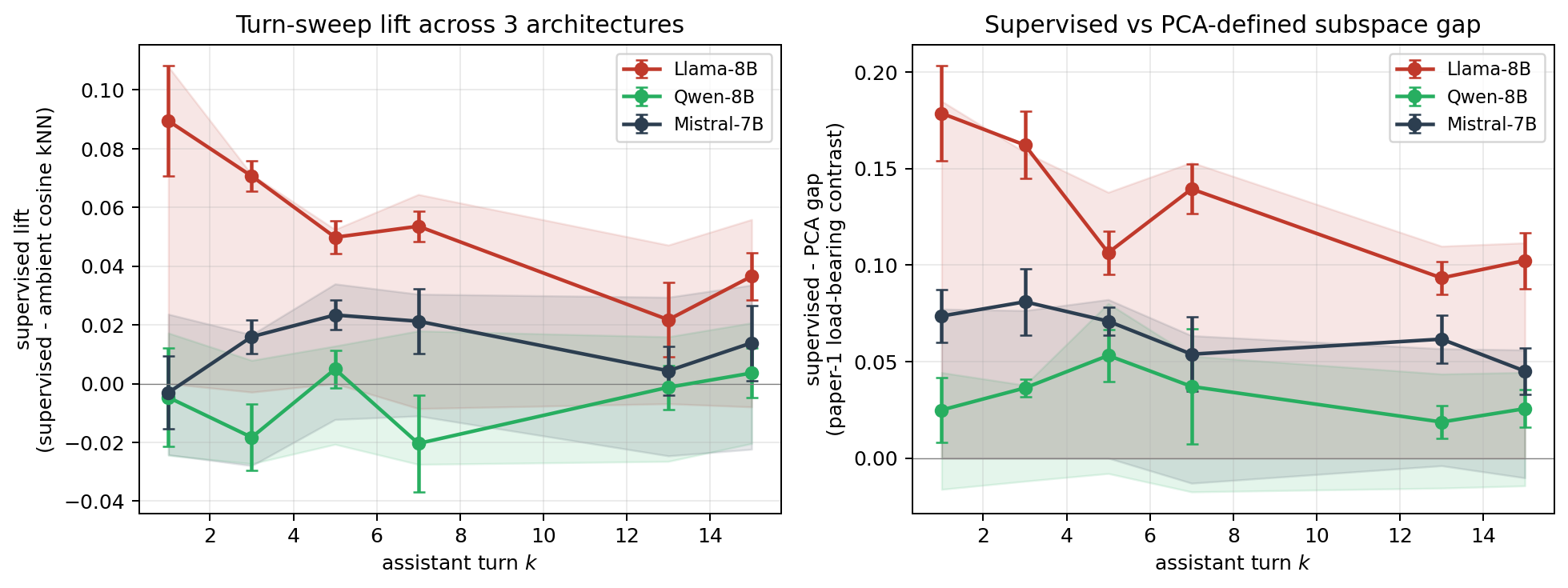}
\caption{Turn-sliced metrics at the headline layer. Left: supervised lift $L_k$ over ambient. Right: supervised$-$PCA gap. Error bars: $5\!\times\!2$ fold-level $\sigma$; faded bands: trajectory-level bootstrap $95\%$ CIs. No systematic accumulation across turns.}
\label{fig:turn-sweep}
\end{figure}

Figure~\ref{fig:turn-sweep} shows the supervised lift $L_k$ and supervised$-$PCA gap $S_k - P_k$ across turns at fixed $N\!=\!600$ per turn. The supervised$-$PCA gap does not systematically increase across turns and is consistent with the cumulative-pool measurement (Table~\ref{tab:headline}).%

Under the pre-registered turn-invariance condition $|L_{T_{15}} - L_{T_1}| \le 0.05$, Qwen ($0.008$) and Mistral ($0.017$) satisfy the criterion, while Llama violates it ($0.053$). We report this directly: one of three models does not satisfy the pre-registered invariance condition. The cumulative-pool supervised$-$PCA gap (Table~\ref{tab:headline}: $0.160$ / $0.142$ / $0.172$) clears the pre-registered $\ge\!0.05$ separation threshold for all three architectures. Taken together, these results indicate that the cosine--linear mismatch does not accumulate over turns and is present throughout the interaction, while the per-model turn-invariance criterion fails for Llama and constrains model-level claims about Llama-specific turn dynamics. Detailed per-turn breakdowns are in Appendix~\ref{app:turn-detail}.

\subsection{Subspace axis: support invariant, within-subspace separability declines for two of three models}
\label{sec:res-subspace}

\begin{figure}[!htbp]
\centering
\includegraphics[width=0.95\linewidth]{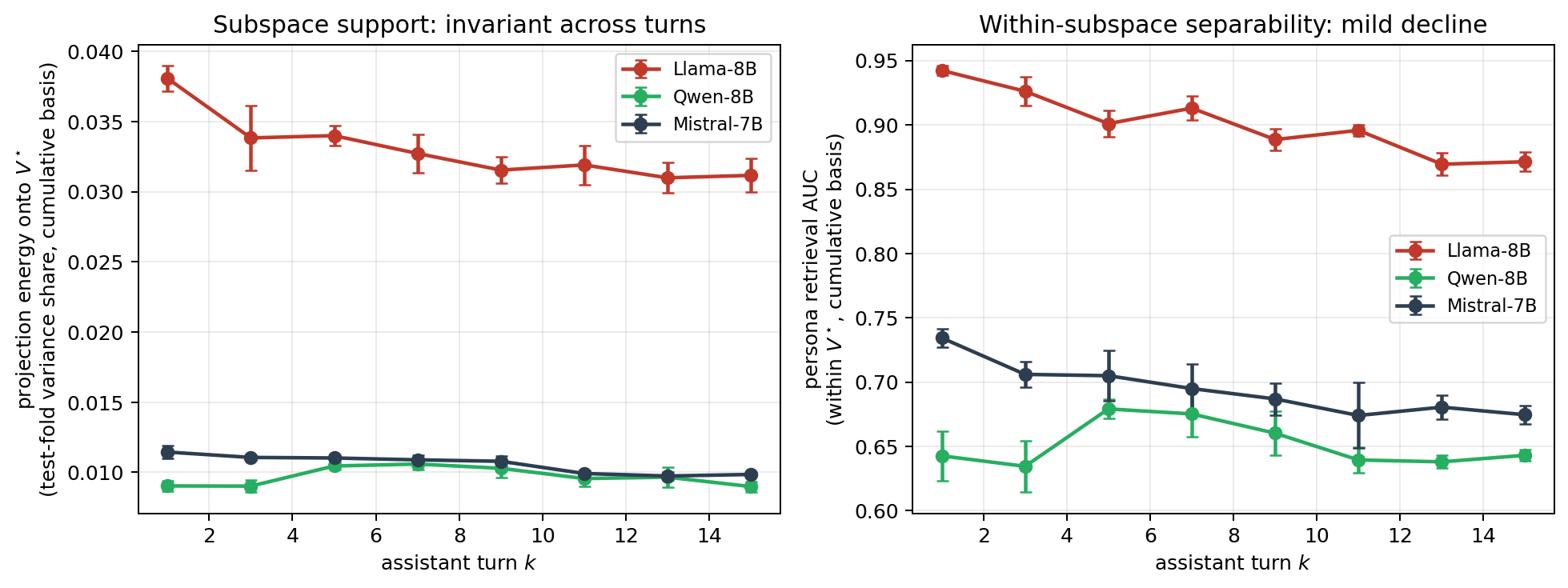}
\caption{Subspace stability across turns under cumulative-pool-anchored $V^\star$. Left: projection energy stable (max-minus-min $\le 0.7$ pp absolute; Llama $\sim 18\%$ relative). Right: within-$V^\star$ persona retrieval AUC, monotone $6$--$7$ pp decline ($T_1\!\to\!T_{15}$) for Llama / Mistral; Qwen-3-8B essentially flat.}
\label{fig:subspace-stability}
\end{figure}

Figure~\ref{fig:subspace-stability} reports projection energy onto the cumulative-pool-anchored basis $V^\star$ (\S\ref{sec:methods-cumulative}) and within-subspace persona retrieval AUC across turns.

Projection energy onto $V^\star$ is invariant across turns at the absolute level (max-minus-min $\le 0.0071$ on all three models, well below the pre-registered $0.05$ threshold), though the Llama trend is monotone and corresponds to roughly an $18\%$ relative decline over the operating range. The persona-relevant subspace support thus does not move appreciably as a fraction of total energy. Within-subspace persona retrieval AUC declines for two of three models ($T_1\!\to\!T_{15}$: $0.942\!\to\!0.871$ Llama, $\Delta=-0.072$; $0.734\!\to\!0.675$ Mistral, $\Delta=-0.060$; Qwen flat at $0.643$, $\Delta\approx 0$). \textbf{Two of three models violate the pre-registered within-subspace AUC invariance threshold $|\Delta| \le 0.03$.} We report this as a pre-registered failure: the AUC invariance criterion is satisfied only for Qwen at the per-turn level, while Llama and Mistral fail. The decline thus occurs within a stable projected subspace rather than from subspace drift---a partial decoupling between representational support and within-subspace separability, which we treat as an empirical constraint rather than evidence of a distinct mechanism (Appendix~\ref{app:subspace-detail}).

\subsection{Axis-decomposition perspective: no consistent cross-architecture pattern}
\label{sec:res-axis}

Per-axis turn deltas remain within $|\Delta|\!\le\!0.07$ across all twelve $(\text{axis}, \text{architecture})$ cells with mixed signs (full table and per-axis turn-sweep figure in Appendix~\ref{app:per-axis}). We find no evidence that the four predefined persona attribute axes (formality, expertise, emotion, domain) exhibit consistent turn-dependent structure in representation space under the tested settings. This statement applies only to the evaluated axes and does not extend to alternative learned or task-specific latent factorizations; our results are consistent with non-axis-aligned structure, but we do not claim identifiability of latent factors.

\FloatBarrier
\section{Discussion}
\label{sec:discussion}

\subsection{Representation geometry vs.\ similarity metrics}

The supervised-vs-PCA dissociation we observe is the empirical signature of a long-standing distinction: the discriminative subspace recovered by a label-trained linear probe and the generative variance subspace recovered by unsupervised PCA do not in general coincide. Ambient cosine similarity, by treating all dimensions equally, inherits the geometry of the latter rather than the former. This pattern is consistent with a broader interpretation: in the regime we study, cosine similarity may fail to reflect task-aligned structure when that structure is not aligned with variance-dominant directions of the ambient representation. We do not establish this as a general property; broader regime coverage is required to determine whether the dialogue-conditioned setting we study is a representative instance.

\subsection{Boundary conditions and constraints}

\paragraph{Interpretive role of the multi-axis analyses.} The turn, subspace, and axis analyses are designed to test specific alternative explanations of the headline cosine--linear gap. If the gap had increased with turn count, it would suggest an accumulation mechanism tied to dialogue progression; if the task-aligned subspace had drifted across turns, it would indicate a moving representation basis rather than a stable one; if a single attribute axis had dominated the cross-architecture pattern, it would suggest axis-specific structure rather than a regime-level effect. None of these patterns are observed at the cumulative-pool measurement: the gap does not systematically increase, the subspace support remains stable in absolute terms, and no single axis exhibits consistent cross-architecture turn dynamics. We treat these negative findings as constraints on the class of plausible explanations for the cosine--linear mismatch, not as standalone phenomena.

Across turns, we find no systematic accumulation, while one model (Llama, $|L_{T_{15}}-L_{T_1}|=0.053$) violates the pre-registered turn-invariance threshold of $0.05$, constraining model-level claims about Llama-specific turn dynamics. Projection energy into the cumulative-pool-anchored subspace is invariant across turns at the absolute level, while within-subspace separability declines for two of three models in violation of the pre-registered $|\Delta\text{AUC}| \le 0.03$ threshold (numerics in \S\ref{sec:res-subspace})---a partial decoupling between representational support and discriminative strength, treated as an empirical constraint rather than a mechanism. Across four predefined persona attribute axes, we observe no consistent turn-dependent pattern; this result applies to the evaluated axes and does not extend to alternative learned latent factorizations.

\subsection{Positioning and methodological recommendation}

We focus on representation-level geometry rather than behavioral outputs and make no causal claim linking representation structure to behavioral phenomena such as persona collapse \citep{chameleon2026,stable2026,heterogeneity2026}. Our results refine the persona-vector and CAA paradigm \citep{panickssery2024caa,anthropic2025personavectors,arditi2024refusal}: the recoverable persona signal is not reducible to a single direction (lift saturates near $k\!\approx\!10$) and is not visible to ambient cosine. The methodological recommendation that follows is summarized as a four-item diagnostic checklist in \S\ref{sec:conclusion}.

\subsection{Limitations}
\label{sec:limitations}

We study three $7$--$8$B chat-tuned model families and a single persona corpus with four predefined attribute axes, capturing hidden states at the last prefill token under group-by-trajectory CV; generalization to larger scales, other corpora, alternative token positions, autoregressive generation, or learned latent axes is open. Our analysis is correlational at the representation level and does not evaluate downstream behavioral consequences. The dialogue-conditioned regime co-varies several factors (multi-turn rollout, system-prompt conditioning, output-state capture, multi-axis attribute structure); within this bundle we isolate attribute cardinality via the matched-cardinality control (\S\ref{sec:res-regime}) but do not separately isolate the remaining factors (Appendix~\ref{app:control-mapping}). Per-turn slice estimates have reduced sample size relative to the cumulative-pool measurements (\S\ref{sec:methods-cumulative}, Appendix~\ref{app:turn-detail}); we anchor regime-level claims on cumulative-pool numbers and report per-turn slice values without extrapolating slice-level conclusions to the full sample.

\FloatBarrier
\section{Conclusion}
\label{sec:conclusion}

We characterize a setting in which \textbf{cosine similarity systematically fails to reflect linearly accessible structure}: dialogue-conditioned multi-turn persona representations across three $7$--$8$B models. We summarize three empirical constraints: (i) the mismatch does not accumulate across the dialogue (Llama violates the pre-registered turn-invariance criterion); (ii) the supervised subspace's projection support is invariant across turns at the absolute level, while within-subspace separability violates the pre-registered AUC-invariance criterion for Llama and Mistral; and (iii) the phenomenon does not decompose along four predefined persona attribute axes. The cosine--linear gap is not a representational failure but a limitation of how similarity-based metrics read structure. \textbf{Practical diagnostic checklist.} We recommend reporting (i) linear-probe AUC, (ii) ambient cosine kNN AUC, (iii) supervised-subspace cosine kNN AUC, and (iv) a matched-rank PCA-subspace cosine kNN AUC; the gap (ii)$-$(iii) and the contrast (iii)$-$(iv) are diagnostic of the structural pattern.

\FloatBarrier

\bibliographystyle{plainnat}
\bibliography{refs}

\appendix

\section{Persona corpus details}
\label{app:personas}

The $30$ personas are manually constructed character profiles spanning four predefined axes (formality $2$-class; expertise $2$-class; emotion $5$-class; domain $5$-class). Rather than sampling from the full $2 \times 2 \times 5 \times 5$ cross-product, we organize personas around an emotion--domain design and vary formality and expertise across instances. Each persona is specified by a system prompt and a short character backstory.

The corpus is designed for diversity rather than balance: we aim to cover a broad range of stylistic and semantic configurations without enforcing uniform marginal or joint distributions over axes. Persona selection was manual and was not optimized for probe separability or downstream evaluation metrics.

To reduce the risk that probes rely on superficial stylistic cues (e.g., tone or formality) rather than semantic persona attributes, we additionally annotate each persona with an auxiliary grouping variable (\texttt{adversarial\_group} A--F and a singleton set), introduced to support post-hoc checks for spurious-feature reliance in linear probes (e.g., whether classification is driven by surface tone similarity rather than persona identity). This grouping is not used in training or evaluation splits, and all main results reported in \S\ref{sec:results} are computed independently of it.

Five seed topics span casual hobby, technical instruction, emotional support, financial advice, and informal troubleshooting. Per (persona, topic) cell, $4$ trajectories are rolled out at $15$ assistant turns each. Prompt and rollout details, including the persona-fixing system prompt template and the user-turn templates, will be released upon publication.

\section{Detailed controls and robustness checks}
\label{app:controls}

\paragraph{L2 normalization (implementation sanity check).} Per-sample L2 normalization is algebraically a no-op for cosine similarity, which is already scale-invariant ($\cos(x,y) = \langle x/\|x\|, y/\|y\| \rangle$). We verified that AUC changes by less than $0.005$ under explicit pre-normalization, confirming our cosine-kNN implementation respects this scale invariance; we do not list this as an independent anisotropy control.

\paragraph{Mean centering plus L2 normalization.} Train-fitted global-mean centering followed by L2 normalization changes ambient cosine kNN AUC by less than $0.005$.

\paragraph{All-but-the-top projection.} At $k\!\in\!\{1, 2, 3, 5, 10, 25\}$ removed top-PCs, ambient kNN AUC remains essentially unchanged or marginally lower. Supervised lift is preserved across all $k$.

\paragraph{Euclidean kNN sanity check.} Euclidean kNN with the same activations gives AUC within $0.02$ of cosine kNN at the headline layer for all three models. The regime contrast and supervised lift are preserved under the alternative metric.

\paragraph{Layer sweep.} The cosine--linear gap is preserved across all six probed layers per model; the supervised lift is monotone increasing with layer depth on Llama and Mistral and approximately flat on Qwen.

\paragraph{$k_\text{knn}$ sweep.} The lift trend across turns is preserved for $k_\text{knn}\!\in\!\{5, 10, 20\}$, ruling out kNN density confounds.

\paragraph{Norm fingerprints.} Mean activation norm at the headline layer: $39.6$ (Llama), $783.2$ (Qwen), $22.7$ (Mistral). Mean pairwise cosine over $500$ random pairs: $0.346$ / $0.618$ / $0.205$. Std/mean ratio of activation norms: $0.089$ / $0.092$ / $0.108$.

\paragraph{Per-turn norm fingerprint.} For each turn $k\!\in\!\{1, 3, 5, 7, 9, 11, 13, 15\}$, mean activation norm and mean pairwise cosine show no monotonic trend that could plausibly drive the lift findings; full per-turn table will be included in the released artifact (\texttt{turn\_sweep\_fingerprint.csv}).

\section{Concept-erasure differentiation}
\label{app:erasure}

A line of work uses supervised subspace projection to \emph{remove} attribute information from representations, with the goal of debiasing or causal-attribute erasure: \citet{bolukbasi2016man} for gender debiasing in word embeddings, \citet{ravfogel2020null} for iterative null-space removal, and \citet{belrose2023leace} for least-squares concept erasure. Our use of supervised subspaces is mechanically related but conceptually inverted: we project \emph{onto} the discriminative subspace rather than removing it, and the goal is measurement (does ambient similarity capture the same structure linear probes do?) rather than intervention (remove an attribute).

The load-bearing distinction in our work is the supervised-vs-PCA contrast. None of the prior concept-erasure methods runs a matched-dimensionality PCA control. Without that control, supervised lifts and supervised projections cannot be distinguished from any low-dimensional projection that incidentally reduces the curse of dimensionality on cosine kNN. The $0$-or-negative PCA lift in our experiments is what licenses the claim that the recovered structure is task-aligned, not variance-aligned.

We do not modify representations; we measure the metric distortion that supervised projection corrects.

\section{Control design and invariance mapping}
\label{app:control-mapping}

The paper makes no claim that the dialogue-conditioned regime is factorized into independent variables; rather, it studies a co-instantiated inference setting, and all controls are designed to test invariance across that joint distribution rather than its components. Each control in the main text targets a specific aspect of the joint distribution:

\begin{itemize}
\item \textbf{Matched-cardinality control} (Appendix~\ref{app:methods-peraxis}, \S\ref{sec:res-regime}): tests whether label cardinality alone accounts for the regime contrast under matched probe capacity and subspace rank. Compares $5$-class persona-emotion (dialogue-conditioned regime) with $5$-class SST-5 (single-sentence regime) at fixed $k_\text{sub}\!=\!\max(\text{n\_classes}-1, 1)$.%
\item \textbf{Per-turn analysis} (Appendix~\ref{app:methods-turn}): tests invariance of the regime-level separation over interaction length. Holds sample size constant across turns ($N\!=\!600$ per turn) and sweeps $k_\text{knn}\!\in\!\{5, 10, 20\}$ at every turn as a density-stability check.
\item \textbf{Cumulative-anchored basis} (\S\ref{sec:methods-cumulative}): tests invariance of subspace support across turns by fitting the supervised basis $V^\star$ on the cumulative pool of all turn activations and projecting per-turn test data onto this fixed basis. Decomposes the per-turn behavior into projection energy (first-order support) and within-subspace AUC (second-order separability).
\end{itemize}

None of these controls assumes or requires factor independence within the regime bundle. We treat the dialogue-conditioned regime as a jointly instantiated operational unit, consistent with prior work on dialogue-conditioned representations where these factors co-vary by design.

\paragraph{Train-fold-only basis fitting rules out label leakage.}
For both projection strategies, the basis (logistic-regression coefficient matrix $W_\text{train}$ for the supervised case, principal components for the PCA case) is fit strictly on the training fold's hidden states and applied unchanged to held-out test trajectories. This rules out the possibility that the supervised lift reflects label leakage between probe training and kNN evaluation: the test fold sees neither the probe's training labels nor any test-side recomputation of the projection basis. Group-by-trajectory cross-validation further ensures that no trajectory contributes to both the projection basis and the kNN evaluation.

\section{Regime contrast robustness}
\label{app:regime-robustness}

\paragraph{Why cardinality alone cannot account for the gap.}
The matched-cardinality regime contrast (\S\ref{sec:res-regime}) holds class count, probe class structure, and projection dimensionality fixed across the two conditions. Under this design, an explanation attributing the supervised lift to class granularity, probe expressivity, or projection dimensionality would predict comparable lift behavior on SST-5 and persona-emotion at matched cardinality. Instead, we observe a divergence: persona-emotion exhibits lifts comparable to the $30$-class setting, whereas SST-5 yields near-zero or negative lift. This pattern is inconsistent with class granularity or probe expressivity as sufficient explanations. We do not interpret the contrast as isolating a unique causal factor; as discussed in \S\ref{sec:res-regime}, the settings differ in sample size, headroom, and within-trajectory dependence. The appropriate conclusion is that the observed pattern is consistent with a regime-level explanation under multiple constraints.

\paragraph{Sample-size robustness.}
The matched-cardinality contrast in \S\ref{sec:res-regime} compares persona-emotion ($\sim\!9{,}000$ pooled-turn activations) with SST-5 ($1{,}000$ sentences), a $\sim 9\times$ asymmetry. To check that the persona-emotion supervised lift does not depend on this, we subsample persona-emotion to $\approx\!1{,}000$ activations (stratified across $5$ emotion classes; mean over $3$ random seeds) and re-run the supervised- and PCA-subspace pipelines at the headline layer ($k_\text{sub}\!=\!5$, group-by-trajectory $2$-fold CV; same hyperparameters as full-$N$). Table~\ref{tab:subsample} reports: supervised lift remains positive on all three architectures (subsampled $+0.081 / +0.046 / +0.077$ vs full-$N$ $+0.096 / +0.130 / +0.154$), as does the supervised$-$PCA gap (subsampled $+0.179 / +0.075 / +0.134$ vs full-$N$ $+0.247 / +0.195 / +0.213$). Subsampled magnitudes attenuate relative to full-$N$, particularly on Qwen (the model with the lowest linear-probe ceiling); the persona-vs-SST-5 sign contrast is preserved on Llama and Mistral, and Qwen's persona-emotion lift ($+0.046$) remains above its SST-5 lift ($+0.013$). Subsampling scripts and seeds ($42, 123, 7$) will be released upon publication.

\begin{table}[!htbp]
\centering
\small
\caption{Sample-size robustness check at headline layer, $k_\text{sub}\!=\!5$, emotion 5-class. Subsampled values are means over 3 random seeds; per-seed numbers will be included in the released CSV.}
\label{tab:subsample}
\begin{tabular}{llcccc}
\toprule
Model & Cond.\ & $N_\text{act}$ & sup.\ lift & sup$-$PCA gap & linear AUC \\
\midrule
Llama-3.1-8B (L$28$)   & full-$N$    & $9{,}000$ & $+0.096$ & $+0.247$ & $0.976$ \\
                       & subsample   & $1{,}000$ & $+0.081$ & $+0.179$ & $0.925$ \\
\midrule
Qwen-3-8B (L$31$)     & full-$N$    & $9{,}000$ & $+0.130$ & $+0.195$ & $0.766$ \\
                       & subsample   & $1{,}000$ & $+0.046$ & $+0.075$ & $0.622$ \\
\midrule
Mistral-7B (L$28$)     & full-$N$    & $9{,}000$ & $+0.154$ & $+0.213$ & $0.815$ \\
                       & subsample   & $1{,}000$ & $+0.077$ & $+0.134$ & $0.697$ \\
\bottomrule
\end{tabular}
\end{table}

\paragraph{Why this is not anisotropy.}
We rule out anisotropy as the explanation for our finding via two controls (full numbers in Appendix~\ref{app:controls}): train-fitted global-mean centering followed by L2 normalization, and all-but-the-top projection at $k = 1, \ldots, 25$. Residual cosine kNN AUC under both controls is within $0.005$ of the unprocessed value. The divergence is therefore not attributable to cone-concentration anisotropy and is not corrected by standard anisotropy-reduction preprocessing. (Per-sample L2 normalization alone is algebraically a no-op for cosine, see Appendix~\ref{app:controls}.)

\paragraph{Why this is not the standard linear-vs-nonlinear probing gap.}
Prior probing work \citep{belinkov2022probing} commonly finds that linear probes outperform distance-based methods. Our contribution goes beyond this observation in two respects. First, the gap is conditional on setting: it appears in dialogue-conditioned multi-axis representations and is absent in single-sentence low-cardinality classification under matched cardinality. Second, the gap closes only after projection onto a task-aligned subspace, not a variance-aligned subspace of matched dimensionality. Without the regime contrast and PCA-matched control, the result would reduce to the standard observation that linear probes are more expressive; with these controls, the evidence instead supports a geometry mismatch between cosine similarity and task-aligned structure in the dialogue-conditioned setting.

\paragraph{Probe regularization sensitivity.}
The linear probe uses fixed $L_2$ regularization $C\!=\!1.0$. To check that the cosine--linear gap is not an artifact of this single choice, we sweep $C \in \{0.1, 1.0, 10.0\}$ at the headline cell (Llama-3.1-8B L$28$, $30$-class persona, $k_\text{sub}\!=\!10$, group-by-trajectory $2$-fold CV). Linear-probe AUC is stable ($0.970$ / $0.972$ / $0.972$), as are ambient cosine kNN ($0.7706$, scale-invariant), supervised-subspace kNN ($0.901$ / $0.909$ / $0.911$), and PCA-subspace kNN ($0.749$ / $0.749$ / $0.749$). The supervised lift ranges from $+0.130$ to $+0.140$ across the sweep (max-min $0.010$), and the supervised$-$PCA gap ranges from $+0.152$ to $+0.162$ (max-min $0.010$). Both quantities remain positive and within $0.02$ across two orders of magnitude in $C$, indicating the gap is not driven by probe-regularization choice. Sweep CSV will be released upon publication.

\section{Methods detail}
\label{app:methods-detail}

\subsection{Per-axis probes (matched-cardinality control)}
\label{app:methods-peraxis}
To distinguish attribute cardinality from representation regime as the explanatory factor, we repeat the analysis per persona axis, with subspace dimensionality $k_\text{sub}\!=\!\max(\text{n\_classes} - 1, 1)$. The $5$-class persona-emotion case is matched in label-cardinality to SST-5 sentiment but remains in the dialogue-conditioned multi-turn regime; comparing it to SST-5 rules out attribute cardinality alone as accounting for the contrast (see \S\ref{sec:res-regime} for the three unmatched factors --- sample size, headroom, within-trajectory dependence --- and the regime-level interpretation under multiple constraints).

\subsection{Sentiment cross-task encoding}
\label{app:methods-sentiment}
We encode $1{,}000$ SST-5 sentences ($200$ per class, stratified random sample from the train split) through all three models using the same hooked layers. Each sentence is presented as a single user-role chat message; the activation at the last prefill token is captured. No system prompt, no multi-turn history, no rollouts. We then run the same four probes under stratified $5$-fold cross-validation.

\subsection{Turn slicing}
\label{app:methods-turn}
For each turn $k\!\in\!\{1, 3, 5, 7, 9, 11, 13, 15\}$, we extract the $600$ activations at that turn position from the multi-turn rollouts and re-run the four metrics. Sample size is held constant across turns. We sweep $k_\text{knn}\!\in\!\{5, 10, 20\}$ at every turn as a density-stability control.

\subsection{Additional controls}
\label{app:methods-controls}
On the same activations: (i) per-sample L2 normalization; (ii) train-fitted global-mean centering followed by L2 normalization; (iii) all-but-the-top projection at $k\!\in\!\{1, 2, 3, 5, 10, 25\}$; (iv) layer sweep across all six probed layers per model; (v) per-layer norm fingerprint (mean, std, min, max activation norm and mean pairwise cosine over $500$ random pairs); (vi) per-turn norm fingerprint for turn-sliced analyses.

\section{Turn-axis detailed results}
\label{app:turn-detail}

For each turn $k\!\in\!\{1, 3, 5, 7, 9, 11, 13, 15\}$ at the headline layer per model, we report the supervised lift $L_k$ and supervised$-$PCA gap $S_k\!-\!P_k$ at $k_\text{sub}\!=\!10$, sample size held constant ($N\!=\!600$ per turn). At $T_1$, the per-turn supervised$-$PCA gap is $0.179$ / $0.025$ / $0.074$ (Llama / Qwen / Mistral); only Llama clears the pre-registered $\ge\!0.05$ separation threshold at the per-turn $T_1$ slice. The per-turn supervised$-$ambient lift at $T_1$ is $+0.089$ / $-0.005$ / $-0.003$. Because per-turn slices have reduced sample size relative to the cumulative pool, we anchor regime-level claims on the cumulative-pool gap (Table~\ref{tab:headline}: $0.160$ / $0.142$ / $0.172$, comfortably clearing $\ge\!0.05$ for all three). Across turns the gap shows no systematic accumulation. The pre-registered turn-invariance condition $|L_{T_{15}}\!-\!L_{T_1}|\!\le\!0.05$ is satisfied for Qwen ($0.008$) and Mistral ($0.017$); Llama violates it ($0.053$), constraining model-level claims about Llama-specific turn dynamics; the cross-architecture cosine-vs-linear gap at the cumulative-pool measurement is unaffected by this Llama-level violation. Trajectory-level bootstrap $95\%$ CIs overlap substantially across turns within each model, and the lift trend is preserved across $k_\text{knn}\!\in\!\{5, 10, 20\}$. Full per-turn fold-level means and standard deviations will be released upon publication (\texttt{paper2\_turn\_sweep/turn\_sweep\_summary.csv}).

\section{Subspace stability detailed results}
\label{app:subspace-detail}

For the cumulative-pool-anchored basis $V^\star$ defined in \S\ref{sec:methods-cumulative}, we report two complementary measures at each turn $k$: projection energy $\|X_k V^{\star\top}\|_F^2 / \|X_k\|_F^2$ (first-order support) and within-$V^\star$ persona retrieval AUC (second-order separability).

\paragraph{Energy is invariant.} Projection energy onto $V^\star$ is stable across turns: Llama-3.1-8B max-minus-min $0.0071$, Qwen-3-8B $0.0016$, Mistral-7B $0.0017$. All three are well below the pre-registered support-invariance threshold of $0.05$. Test-fold activations at any turn occupy the cumulative-pool subspace to the same proportional extent: the persona-relevant subspace is fully formed at $T_1$ and does not move.

\paragraph{Within-subspace AUC violates the pre-registered invariance criterion for two of three models.} Persona retrieval AUC within $V^\star$ exhibits a monotone or near-monotone decline for Llama-3.1-8B ($0.942$ at $T_1$ $\to$ $0.871$ at $T_{15}$, $\Delta\!=\!-0.072$) and Mistral-7B ($0.734 \to 0.675$, $\Delta\!=\!-0.060$); Qwen-3-8B is essentially flat ($0.643 \to 0.643$, $\Delta\!\approx\!0$). Two of three models violate the pre-registered AUC-invariance threshold of $|\Delta| \le 0.03$, consistent with the main-text framing in \S\ref{sec:res-subspace}.

\paragraph{Interpretation.} Energy is a first-order \emph{support} measure: does the activation occupy the cumulative subspace? The answer is yes, with energy stable to within $0.7$ percentage points. Within-subspace AUC is a second-order \emph{separability} measure: given that the activation lies in the subspace, how clustered are persona classes within it? The pre-reg-violating decline for two of three models (Llama, Mistral) suggests a partial decoupling between stability of representational support and discriminative strength within that support. We treat this as an empirical constraint on representation dynamics rather than evidence of a distinct mechanism. The two findings are complementary: the mismatch we report at the regime level is not due to a lack of linearly decodable structure in the representation; the structure is present in a stable subspace and exhibits a within-subspace decline that violates the pre-registered AUC-invariance criterion on two of three models.

\section{Related-work expanded differentiation}
\label{app:related-detail}

\paragraph{Persona representation and persona-conditioned behavior.}
A growing line of work studies persona conditioning in large language models. Behavioral studies report that persona-conditioned outputs may collapse or homogenize over long interactions \citep{chameleon2026,stable2026,heterogeneity2026}, and other work characterizes persona consistency as a function of probing or steering choices \citep{systematic_persona2026,emergent_misalign2026}. Our analysis is at the representation level rather than the behavior level: we measure whether structure is decodable from hidden states by ambient versus supervised similarity-based metrics, not whether persona-conditioned outputs are behaviorally distinguishable.

\paragraph{Subspace projection methods (EpiPersona).}
EpiPersona \citep{epipersona2026} projects from prompt and context features onto a low-dimensional persona space, a data-driven, correlational mapping. Our supervised subspace $V^\star$ is derived directly from the linear-probe coefficient matrix on internal hidden states, which is representation-derived and decoding-anchored. The two approaches answer different questions: EpiPersona asks whether persona-relevant structure can be inferred from input features; we ask whether such structure is encoded in hidden states and whether it is missed by ambient similarity metrics.

\paragraph{Persona vectors and steering.}
Contrastive Activation Addition (CAA) and persona vectors \citep{panickssery2024caa,anthropic2025personavectors,arditi2024refusal} extract a contrastive direction between positive and negative examples and use it as a steering vector. \citet{tan2024analyzing} shows that steering vectors generalize unevenly across input distributions; \citet{engels2024notall} show that some features are inherently multi-dimensional rather than single-direction. Our analysis is upstream of intervention: we treat the recoverable persona signal as concentrated in a low-dimensional task-aligned subspace, demonstrate that ambient similarity metrics fail to reflect this structure, and leave intervention to follow-up work.

\paragraph{Pragmatic and structural persona encoding.}
Pragmatic Persona \citep{pragmatic2026} reports that persona is structurally encoded in discourse organization rather than purely in lexical surface features, an observation orthogonal to but consistent with the structural-encoding picture our findings support. We make no causal claim about which structural encoding mechanism is responsible for the cosine--linear mismatch we observe; we merely note the alignment in observation.

\paragraph{Concurrent persona-geometry work.}
\citet{wang2025persona} studies persona representations on Qwen2.5-0.5B and proposes a dual-head probing architecture combined with contrastive activation addition for deterministic persona steering, assuming the linear representation hypothesis. Our work is complementary: rather than building a steering apparatus that \emph{assumes} persona directions are linearly recoverable, we test whether ambient cosine geometry is aligned with those directions, and quantify the regime-dependent divergence cross-architecturally on $7$--$8$B-class models. Wang assumes geometry; we test whether a specific geometry (cosine) reflects task structure. The two papers approach persona geometry from orthogonal angles---a positive construction (build a controller) versus a regime-contrast diagnostic (show standard metrics underestimate structure on multi-axis dialogue regimes only). Our regime-contrast finding therefore provides measurement-level support for subspace-based persona analysis: in dialogue-conditioned regimes, projection onto a task-aligned linear subspace---rather than ambient cosine---is the appropriate geometry for similarity-based persona evaluation and steering verification.

\section{Per-axis full numerical results}
\label{app:per-axis}

\begin{figure}[H]
\centering
\includegraphics[width=0.95\linewidth]{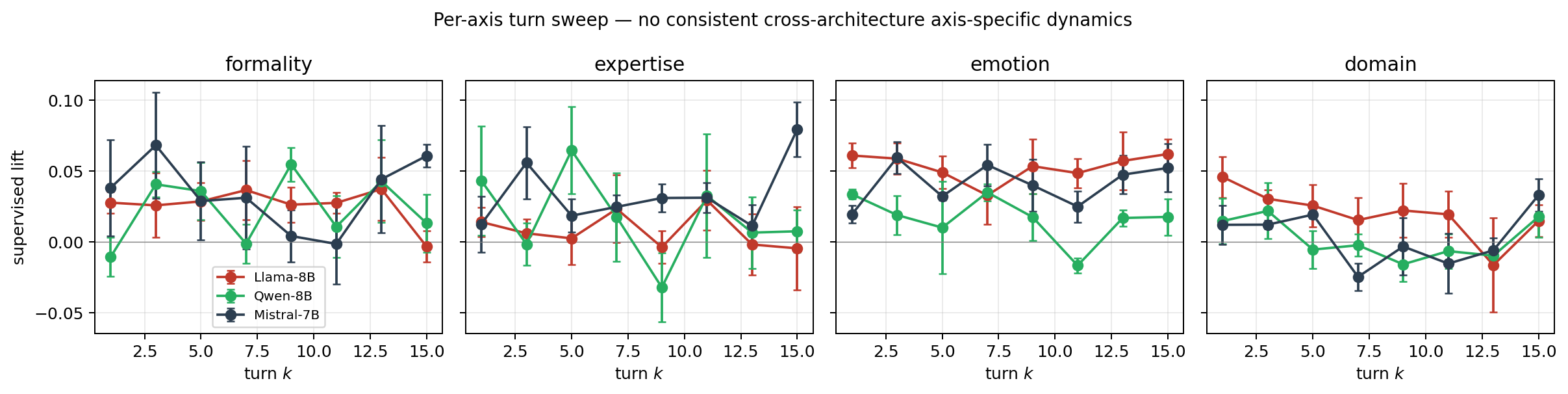}
\caption{Per-axis turn sweep for the four predefined persona axes (formality, expertise, emotion, domain) across three architectures. Per-axis turn deltas $\Delta\!=\!L_{T_{15}}\!-\!L_{T_1}$ remain within $|\Delta|\!\le\!0.07$ across all twelve $(\text{axis}, \text{architecture})$ cells with mixed signs.}
\label{fig:per-axis}
\end{figure}

Table~\ref{tab:per-axis} gives the per-axis turn delta $\Delta_k\!=\!L_{T_{15}}\!-\!L_{T_1}$ for each (axis, architecture) cell, and Figure~\ref{fig:per-axis} shows the per-axis turn sweeps.

\begin{table}[H]
\centering
\small
\caption{Per-axis turn deltas $L_{T_{15}}\!-\!L_{T_1}$ across four predefined persona axes and three architectures. All twelve cells satisfy $|\Delta|\!\le\!0.07$, with mixed signs across architectures.}
\label{tab:per-axis}
\begin{tabular}{lrrr}
\toprule
Axis & Llama-3.1-8B $\Delta$ & Qwen-3-8B $\Delta$ & Mistral-7B $\Delta$ \\
\midrule
formality & $-0.031$ & $+0.024$ & $+0.022$ \\
expertise & $-0.019$ & $-0.036$ & $+0.067$ \\
emotion   & $+0.001$ & $-0.016$ & $+0.033$ \\
domain    & $-0.031$ & $+0.003$ & $+0.021$ \\
\bottomrule
\end{tabular}
\end{table}

\section{Pre-registration}
\label{app:prereg}

\paragraph{Variance reporters.} (1) $5\!\times\!2$ grouped cross-validation fold-level standard deviation, with \texttt{group=trajectory\_id} preserved. (2) Trajectory-level bootstrap $95\%$ confidence intervals ($1{,}000$ resamples drawing trajectories with replacement; bootstrap unit is the trajectory).

\paragraph{Decision rules.}
\begin{itemize}
\item Regime gap: established if supervised lift $\ge\!0.05$ and $S\!-\!P\!\ge\!0.05$ on persona task and not on SST-5.
\item Turn invariance: $|L_{T_{15}}\!-\!L_{T_1}|\!\le\!0.05$.
\item Subspace support invariance: energy max-minus-min $<\!0.05$ per model.
\item Subspace separability invariance: AUC max-minus-min $<\!0.03$ per model.
\end{itemize}

\paragraph{Trigger condition for zero-context micro-pilot.}
A zero-context micro-pilot (capturing activations at the end of the system prompt before any user message) was pre-registered to run only if turn invariance failed (a measurable accumulation curve was observed). The trigger condition was not met and the micro-pilot was not run.

\paragraph{Density-stability check.}
The supervised-subspace kNN was swept over $k_\text{knn}\!\in\!\{5, 10, 20\}$ at every turn. The headline lift trend is preserved across all three values; the lift trend across turns is not driven by density confounds.

\section{Code release manifest}
\label{app:code}

Upon publication, we will release: (1) multi-turn persona pilot data (residual-stream activations at six layers per model, all $15$ turns, $600$ trajectories per model); (2) SST-5 encoded activations under the same protocol; (3) all six probe / subspace / kNN / PCA scripts and per-axis variants; (4) turn-sweep, cumulative-anchored-subspace, and per-axis analysis scripts; (5) figure-rendering scripts; (6) README documenting expected directory structure and dependencies (\texttt{numpy}, \texttt{scikit-learn}, \texttt{matplotlib}, \texttt{pandas}).

\end{document}